\documentclass{article}

\usepackage{algorithm}
\usepackage{algorithmicx}
\usepackage[noend]{algpseudocode}

\algnewcommand\algorithmicassert{\texttt{assert}}
\algnewcommand\Assert[1]{\State{} \algorithmicassert(#1)}

\algnewcommand\algorithmicswitch{\textbf{switch}}
\algnewcommand\algorithmiccase{\textbf{case}}
\algnewcommand\algorithmicdefault{\textbf{default}}
\algdef{SE}[SWITCH]{Switch}{EndSwitch}[1]{\algorithmicswitch\ #1\ \algorithmicdo}{\algorithmicend\ \algorithmicswitch}
\algdef{SE}[CASE]{Case}{EndCase}[1]{\algorithmiccase\ #1}{\algorithmicend\ \algorithmiccase}
\algdef{SE}[DEFAULT]{Default}{EndDefault}[1]{\algorithmicdefault\ #1}{\algorithmicend\ \algorithmicdefault}
\algtext*{EndSwitch}
\algtext*{EndCase}
\algtext*{EndDefault}

\algnewcommand\algorithmicmatch{\textbf{match}}
\algdef{SE}[MATCH]{Match}{EndMatch}[1]{\algorithmicmatch\ #1}{\algorithmicend\ \algorithmicmatch}
\algtext*{EndMatch}

\algnewcommand\algorithmictry{\textbf{try}}
\algnewcommand\algorithmiccatch{\textbf{catch}}
\algblockdefx[Try]{Try}{EndTry}{\algorithmictry}{\algorithmicend\ \algorithmictry}
\algtext*{EndTry}
\algcblockdefx[Catch]{Try}{Catch}{EndTry}[1]{\algorithmiccatch\ #1}{\algorithmicend\ \algorithmictry}%
\algtext*{EndTry}
\MakeRobust{\Call}

\algdef{SE}[QUERY]{Query}{EndQuery}%
   [2]{\textbf{query}\ \textproc{#1}\ifthenelse{\equal{#2}{}}{}{(#2)}}%
   {\algorithmicend\ \algorithmicquery}%
\algtext*{EndQuery}

\def\@algocf@capt@ruled{under}

\usepackage{microtype}
\usepackage{graphicx}
\usepackage{booktabs} 

\usepackage[accepted]{icml2019}

\usepackage{url} 
\usepackage{nicefrac} 
\usepackage{subcaption}
\usepackage{array}
\usepackage{float}
\usepackage{amsmath}
\usepackage{amssymb}
\usepackage{amsfonts}
\usepackage{latexsym}

\newcommand{\eval}{\,\reflectbox{$\leadsto$}\:}

\icmltitlerunning{Nested Reasoning About Autonomous Agents Using Probabilistic Programs}

\begin{document}

\twocolumn[
\icmltitle{Nested Reasoning About Autonomous Agents Using Probabilistic Programs}



\icmlsetsymbol{equal}{*}

\begin{icmlauthorlist}
\icmlauthor{Iris R. Seaman}{neu}
\icmlauthor{Jan-Willem van de Meent}{equal,neu}
\icmlauthor{David Wingate}{equal,byu}
\end{icmlauthorlist}

\icmlaffiliation{neu}{Khoury College of Computer and Information Sciences, Northeastern University, Boston, MA, USA}
\icmlaffiliation{byu}{Department of Computer Science, Brigham Young University, Provo, UT, USA}

\icmlcorrespondingauthor{Jan-Willem van de Meent}{j.vandemeent@northeastern.edu }
\icmlcorrespondingauthor{David Wingate}{wingated@cs.byu.edu }

\icmlkeywords{Machine Learning, ICML}

\vskip 0.3in
]



\printAffiliationsAndNotice{\icmlEqualContribution} 

\begin{abstract}
As autonomous agents become more ubiquitous, they will eventually have to reason about the plans of other agents, which is known as \emph{theory of mind} reasoning. We develop a planning-as-inference framework in which agents perform nested simulation to reason about the behavior of other agents in an online manner. As a concrete application of this framework, we use probabilistic programs to model a high-uncertainty variant of pursuit-evasion games in which an agent must make inferences about the other agents' plans to craft counter-plans. Our probabilistic programs incorporate a variety of complex primitives such as field-of-view calculations and path planners, which enable us to model quasi-realistic scenarios in a computationally tractable manner. We perform extensive experimental evaluations which establish a variety of rational behaviors and quantify how allocating computation across levels of nesting affects the variance of our estimators.





%

\end{abstract}

\section{Introduction}

An autonomous agent that interacts with other agents needs to do more than simply perceive and respond to their environment. 
Eventually agents will need to reason about all of
the complexities inherent in the real world, including 
the beliefs, intents and desires of other intentional agents.  This is
known as \textit{theory of mind}, and is indispensable if we hope to
one day create agents capable of empathy, ``reading between the
lines,'' and interacting with humans as peers.

In this paper, we explore how theory of mind can be implemented on high-uncertainty pursuit-evasion games using nested simulations in the form of probabilistic programs. Contrary to classic pursuit-evasion problems, which are minimax games with fully-observable environments, we develop partially observable variants of the which have a high obstacle count, limited field of view, and noisy trajectory planning. We develop a domain in which a \emph{chaser} agent must reason about possible locations of a \emph{runner} agent, who seeks to avoid detection. The runner's intended start location, goal location, and likely path to the goal are initially unknown to the chaser. The runner knows the current location of the chaser, but not the chaser's future trajectory. This results in a setting where agents must reason about the reasoning of other agents under both state and outcome uncertainty.


We perform online planning in this domain using planning-as-inference approach \cite{toussaint06}. We formulate the chaser and the runner models as nested probabilistic programs that are conditioned according to the desired behavior of the respective agent. The model of the chaser is conditioned to \emph{maximize} the likelihood of detection, and the runner is conditioned to \emph{minimize} likelihood of detection. At each point of time, the chaser imagines possible future trajectories, along with possible runner trajectories, and selects a move that has a high probability of detection. 

Formulating our models as probabilistic programs also make it possible to incorporate semi-realistic deterministic primitives such as path planners and visibility graphs. 
Moreover, an advantage of our planning-as-inference approach is that we can employ nested importance sampling methods \cite{naesseth2015nested} for probabilistic program inference to perform recursive reasoning. This enables us to implement tractable inference in a partially-observable multi-agent domain with continuous actions. To our knowledge, this work is the first to adapt nested importance sampling methods to online planning in multi-agent systems.

We evaluate a range of scenarios to demonstrate that nested Bayesian reasoning leads to rational behaviors in which agents maximize the relative utility at each level in the model. We evaluate the effect of model complexity on runner detection rates relative to basic models. Finally we perform extensive experiments to quantify the effect of allocation of computational budget across levels in the model on the variance of the estimated expected utility.

\begin{figure*}
\begin{center}
\centerline{\includegraphics[width=0.9\textwidth]{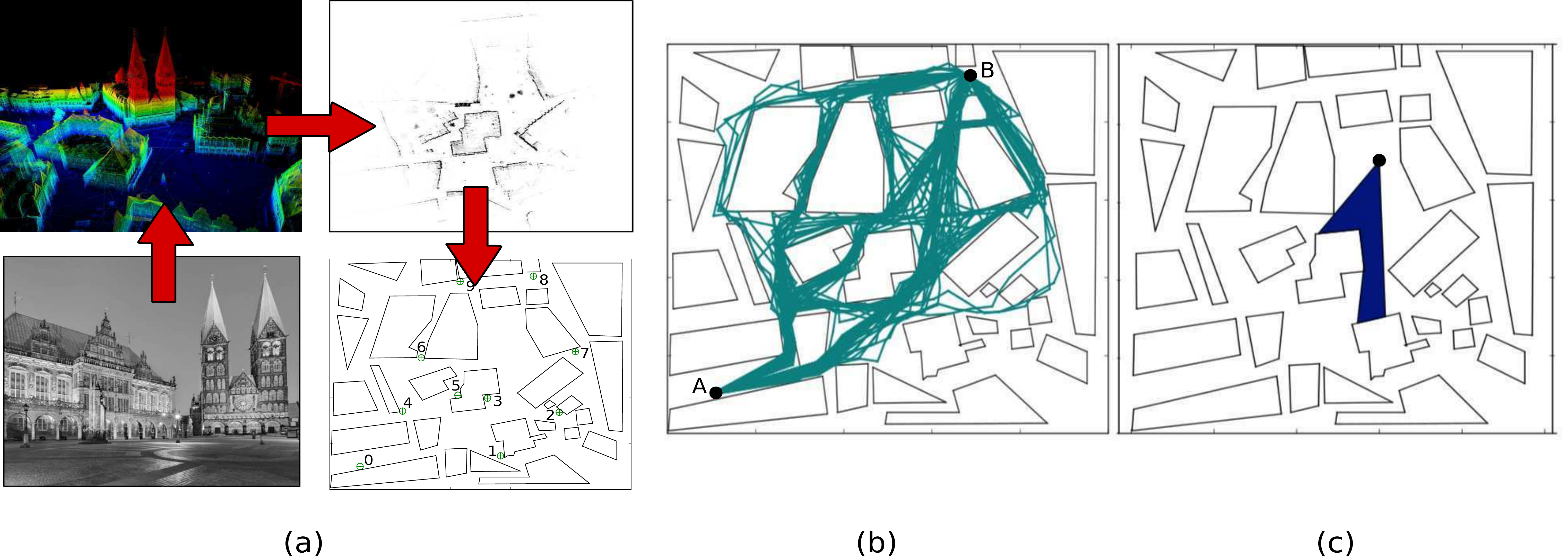}}
\caption{(a) We generate a coarse polygonal city map from point cloud data of the city of Bremen, Germany. (b) Visual distribution over paths a runner may take
modeled with Random-Exploring Random Trees, RRTs, from points A to B. (c) a 45$^{\circ}$ isovist, or range of sight, of the
chaser.  The isovist is properly blocked by buildings.}
\label{fig:rrt}
\end{center}
\vspace{-1.0em}
\end{figure*} 

\section{Background}




\subsection{Theory of Mind}

Human children develop theory of mind during their
early years, generally between the ages of three and
six
\cite{wellman1990child,chater2006probabilistic}. \citet{bello2006developmental}
explore this phenomenon with a computational model that suggests that the underlying cognitive shifts required for the development of theory of mind may be smaller than previously supposed. 
\citet{goodman2006intuitive} present a formal model that
attempts to account for false belief in
children, and later take the innovative approach of
linking inference with causal reasoning \cite{goodman2009cause}. Additionally, the same group explores language as a type of social
cognition \cite{goodman2013knowledge}.

The development of theory of mind in machines leads naturally to interaction with their human counterparts. \citet{awais2010human}, \citet{fern2007decision}, and \citet{nguyen2012capir} investigate collaboration between humans and robots in which the robot must determine the human's (unobservable) goal. In a complementary line of research, \citet{sadigh2016information} explore the idea of active information, in which the agent's own behaviors become a tool for identifying a human's internal state. Fully-developed theory of mind requires the possibility of nested beliefs. \citet{koller1997effective} present an inference algorithm for recursive stochastic programs. \citet{frith2005theory} argue that theory of mind can be modeled using probabilistic programming, and demonstrate examples of nested conditioning with the probabilistic programming language, Church. \citet{zettlemoyer2009multi} address filtering in environments with many agents and infinitely nested beliefs. 

To our knowledge, our work is the first to model nested reasoning about agents in a time-dependent manner. Prior work by \citet{baker2009action} develops a Bayesian framework for reasoning about preferences of individual agents based on observed time-dependent trajectories. Our work differs in that our environment is not discretized into a grid world, and as such represents a continuous action space. Work by \citet{stuhlmuller2014reasoning} employed probabilistic programs to model nested reasoning about other agents. 
Relative to this work, our work differs in that agents update and \textit{act upon} their beliefs of other agents in a time-dependent manner, whereas the work by \citet{stuhlmuller2014reasoning} considers problems for a single decision.

\subsection{Probabilistic Program Inference}

To represent our generative model cleanly and to perform inference in it, we employ the tools of probabilistic programming \cite{vandemeent2018introduction}. This allows us to define probabilistic models that incorporate control flow, libraries of deterministic primitives, and data structures. 
A probabilistic program is a procedural model that, when run unconditionally, yields a sample from a prior distribution. Running probabilistic programs forward is fast and only limited by the native speed of the interpreter for the language.

Inference in probabilistic programming involves reasoning about a target distribution that is conditioned by a likelihood, or more generally a notion of utility \cite{vandemeent2018introduction}.
Inference for probabilistic programs is difficult because of the flexibility that probabilistic programming languages provide: an inference algorithm must behave reasonably for any program a user wishes to write. Many probabilistic programming systems rely on Monte Carlo methods due to their generality \cite{goodman08,milch05,pfeffer01,standevelopmentteam2014stan,venture}. Methods based on importance sampling and sequential Monte Carlo (SMC) have become popular \cite{murray2013,todeschini2014biips,wood-aistats-2014,goodman2014dippl,ge2016turing}, due to their simplicity and compositionality \cite{naesseth2015nested}.

For our purposes, the most important feature of probabilistic programming languages is that they allow us to freely mix deterministic and stochastic elements, resulting in tremendous modeling flexibility.  This makes it relatively easy to (for example) describe distributions over Rapidly-Exploring Random Tree (RRTs), isovists (field-of-view calculations), or even distributions that involve optimization problems as a subcomponent of the distribution.

\section{Simulation Primitives}

Although probabilistic programming has previously been used to model theory of mind \cite{stuhlmuller2014reasoning}, past implementations have thus far considered relatively simple problems involving a small number of decisions. In this paper, we not only model a setting in which agents must reason about future events, but also do so in a manner that involves reasoning about properties of the physical world. To enable this type of reasoning, we will employ a number of semi-realistic simulation primitives.

\textbf{The environment.}  To search for and intercept the runner, the
chaser requires a representation of the world that allows reasoning
about starting locations, goals, plans, movement and visibility. We
use a polygonal model designed around a known, fixed map of the city of
Bremen, Germany \cite{BremenPointCloud}, shown in Fig.~\ref{fig:rrt} (a).

\textbf{Path planning and trajectory optimization.}  We model paths
using a RRT
\cite{lavalle1998rapidly}, a randomized path planning algorithm
designed to handle nonholonomic constraints and high degrees of
freedom.  We leverage the random nature of the RRT to describe an
entire distribution over possible paths: each generated RRT path can
be viewed as a sample from the distribution of possible paths taken by
a runner (see Fig.~\ref{fig:rrt} (b)). RRTs naturally consider short
paths as well as long paths to the goal location. To foreshadow a bit,
note that because we will be performing inference over RRTs
conditioned on not being detected, the runner will naturally tend to
use paths that minimize the chance of detection, which are often, but
not always, the shortest and most direct.  Our RRTs are
refined using a trajectory optimizer to eliminate bumps
and wiggles.

\textbf{Visibility and detection.}  Detection of the runner by the
chaser is modeled using an isovist, a polygon
representation of the chaser's current range of sight
\cite{isovist79,morariu2007human}. Given a map, chaser location, and
runner location, the isovist determines the likelihood
that the runner was detected. Although an isovist usually uses a 360
degree view to describe all possible points of sight to the chaser, we
limit the range of sight to 45
degrees, and add direction to the chaser's sight as seen in
Fig.~\ref{fig:rrt} (c). The direction of the chaser's line of sight is determined by the imagined location of the runner.


\section{The Chaser-Runner Model}
\label{sec:tcrm}

To model theory of mind, we will develop a nested probabilistic program in which a chaser plans a trajectory by maximizing the probability of interception relative to imagined runner trajectories. The model for runner trajectories, in turn, assumes that the runner imagines chaser trajectories and avoids paths with a high probability of interception. 

Our model has four levels: the \textbf{episode model} samples a sequence of moves by the chaser. Each move is sampled from the \textbf{outermost model}, which describes the
beliefs of the chaser about the expected utility of moves. This model compares future chaser trajectories to imagined runner trajectories and assigns higher probability to trajectories in which detection  is likely. The runner trajectories are in turn sampled from the  \textbf{middlemost model}, which minimizes detection probability based on imagined chaser trajectories that are sampled from the \textbf{innermost model}.
These models work in tandem to create nuanced inferences about
where the chaser believes the runner might be, and what 
counter-plan will maximize the probability of detection.

Algorithm~\ref{alg:chaser-runner-programs} shows pseudo-code for the Chaser-Runner model, formulated as nested probabilistic programs, which we refer to as queries. Together, these programs define a planning-as-inference problem \cite{toussaint06} in which queries generate weighted samples, resulting in a nested importance sampling  \cite{naesseth2015nested} scheme that we describe in more detail below.

\textbf{The episode model} initializes the location of the chaser to a specified start location $x^\textsc{c}_1 = x^\textsc{c}_\textsc{start}$. For time points $t=2 \ldots T$, the model samples weighted partial trajectories $(x^\textsc{c}_{1:t}, w_t)$ from the outermost \textsc{chaser} model. After the final iteration, the model returns full trajectories $x^\textsc{c}_{1:T}$.

\textbf{The outermost model} describes the chaser's plan for trajectories, given the chaser's belief about possible runner trajectories. The chaser selects a goal location $x^\textsc{c}_\textsc{goal}$ at random and uses the RRT planner to sample a possible future trajectory $x^\textsc{c}_{t:T}$. Note that this trajectory is random, owing to the stochastic nature of the RRT algorithm. In order to evaluate the utility of this trajectory, the chaser imagines a possible runner trajectory by sampling from the middlemost \textsc{runner} model. The chaser then evaluates the utility of the trajectory by using an isovist representation to determine the number of time points $T^\textsc{c}_\textsc{visible}$ during which the runner is visible to the chaser. The chaser then conditions the sampled trajectories by defining a weight $w^\textsc{c} = \exp( \alpha \, T^\textsc{c}_\textsc{visible})$. As we will discuss below, this corresponds to assigning a utility proportional to $T^\textsc{c}_\textsc{visible}$ in a planning-as-inference formulation. The model discards most of the imagined future trajectory, keeping only the next time point $x^\textsc{c}_t$, and returns the partial trajectory $x^\textsc{c}_{1:t}$, together with a weight $w^\textsc{c} \cdot w^\textsc{r}$ that reflects the utility of the chaser and the runner.

\begin{algorithm}[!t]
\begin{algorithmic}[1]
\Query{episode}{$x^\textsc{c}_\textsc{start}$}
    \Comment{Episode model}
    \For{$k$ \textbf{in} $1 \ldots K$}
    \State $x^{\textsc{c},k}_1 = x^\textsc{c}_\textsc{start}$
    \EndFor
    \For{$t$ \textbf{in} $2 \ldots T$}
    \For{$k$ \textbf{in} $1 \ldots K$}
    \State $x^{\textsc{c},k}_{1:t}, w^k_t \eval \textsc{chaser}(x^{\textsc{c},k}_{1:t-1})$ \EndFor
    \For{$k$ \textbf{in} $1 \ldots K$}
        \State $a \sim \text{Categorical}\left(\frac{w^1_t}{\sum_k w^k_t}, \dots, \frac{w^K_t}{\sum_k w^k_t} \right)$
        \State $x^{\textsc{c},k}_{1:t}, w^k_t = x^{\textsc{c},a}_{1:t}, \frac{1}{K} \sum_k w^k_t$
    \EndFor
    \EndFor
    \State \Return $(x^{\textsc{c},1}_{1:T}, w^1_T), \dots, (x^{\textsc{c},K}_{1:T}, w^K_T)$
\EndQuery
\Query{chaser}{$x^\textsc{c}_{1:t-1}$}
    \Comment{Outer Model}
    \State $x^\textsc{c}_\textsc{goal} \sim \text{Uniform}(\{x_\textsc{a},\ldots,x_\textsc{j}\})$
    \State $x^\textsc{c}_{t:T} \sim \Call{rrt-plan}{x^\textsc{c}_{t-1}, x^\textsc{c}_\textsc{goal}}$
    \For{$l$ \textbf{in} $1 \ldots L$}
      \State $x^{\textsc{r},l}_{t:T}, w^{\textsc{r},l} \eval \textsc{runner}(x^\textsc{c}_{1:t-1})$
      \State $T^{\textsc{c},l}_\textsc{visible} = \Call{time-visible}{x^{\textsc{r},l}_{t:T}, x^\textsc{c}_{t:T}}$
      \State $w^{\textsc{r},l} = \exp(\alpha \; T^{\textsc{c},l}_\textsc{visible})$
    \EndFor
    \State \Return $x^\textsc{c}_{1:t}, w^\textsc{c} \cdot \left(\frac{1}{L} \sum_{l} w^{\textsc{r},l}\right)$
\EndQuery
\Query{runner}{$x^\textsc{c}_{1:t-1}$} \Comment{Middle Model}
    \State $x^\textsc{r}_\textsc{start} \sim \text{Uniform}(\{x_\textsc{a},\ldots,x_\textsc{j}\})$
    \State $x^\textsc{r}_\textsc{goal} \sim \text{Uniform}(\{x_\textsc{a},\ldots,x_\textsc{j}\})$
    \State $x^\textsc{r}_{1:T} \sim \Call{rrt-plan}{x^\textsc{r}_\textsc{start}, x^\textsc{r}_\textsc{goal}}$
    \State $\tilde{x}^\textsc{c}_{t:T}, \tilde{w}^\textsc{c} \eval \textsc{naive-chaser}(x^\textsc{c}_{t-1})$
    \State $T^\textsc{r}_\textsc{visible} = \Call{time-visible}{x^\textsc{r}_{1:T}, \{x^\textsc{c}_{1:t-1}, \tilde{x}^\textsc{c}_{t:T}\}}$
    \State $w^\textsc{r} = \exp(-\alpha \; T^\textsc{r}_\textsc{visible})$
    \State \Return $x^\textsc{r}_{t:T},  w^\textsc{r} \cdot \tilde{w}^\textsc{c}$
\EndQuery
\Query{naive-chaser}{$x^\textsc{c}_{t-1}$} \Comment{Inner Model}
    \State $\tilde{x}^\textsc{c}_\textsc{goal} \sim \text{Uniform}(\{x_\textsc{a},\ldots,x_\textsc{j}\})$
    \State $\tilde{x}^\textsc{c}_{t:T} \sim \Call{rrt-plan}{x^\textsc{c}_{t-1}, \tilde{x}^\textsc{c}_\textsc{goal}}$ 
    \State \Return $\tilde{x}^\textsc{c}_{t:T}, 1$
\EndQuery
\end{algorithmic}

\caption{
\label{alg:chaser-runner-programs}
Schematic representation of the Chaser-Runner model. The \textsc{episode} model performs SMC in which moves are sampled from a nested \textsc{chaser} model, which in turn simulates runner trajectories from a second nested \textsc{runner} model. The \textsc{chaser} model is conditioned to \emph{maximize} the probability of future detections, whereas the \textsc{runner} model is conditioned \emph{minimize} both past and future detections. At each time $t$, we propose $K$ future trajectories for the chaser and $K \times L$ trajectories for the runner.}
\end{algorithm}

\textbf{The middlemost model} describes the chaser's reasoning about possible runner trajectories. 
We assume that the chaser models a worst-case scenario where the runner is aware of the chaser's location. This could be, for example, because the runner uses a police scanner to listen in on the chaser's reported location.
Moreover, we assume that at any point in time, the episode only continues when the chaser has not yet detected the runner. Finally, we assume that the runner will seek to avoid detection by imagining a chaser trajectory, and then selecting a trajectory that will not intersect that of the chaser. We implement these assumptions in the probabilistic program as follows. The runner model first selects a start location $x^\textsc{r}_\textsc{start}$ and goal location $x^\textsc{r}_\textsc{goal}$ at random, and then samples a random trajectory $x^\textsc{r}_{1:T}$ using the RRT planner. The runner then imagines a future chaser trajectory by selecting a goal location $\tilde{x}^\textsc{c}_\textsc{goal}$ at random and sampling $\tilde{x}^\textsc{c}_{t:T}$ from the innermost model. We then condition this sample by computing the total time of visibility $T^\textsc{r}_\textsc{visible}$, based on both the known past trajectory $x^\textsc{c}_{1:t-1}$ and the imagined future trajectory $\tilde{x}^\textsc{c}_{t:T}$ of the chaser. Finally, we assign a weight $w^\textsc{r} = \exp(-\alpha T^\textsc{r}_\textsc{visible})$, which corresponds to a negative utility (i.e.~a cost) proportional to $T^\textsc{r}_\textsc{visible}$ in the planning-as-inference formulation.

\textbf{The innermost model} describes future chaser trajectories imagined by the runner. This model is the simplest of all the models in our nested formulation. Given the previous location $x^\textsc{c}_{t-1}$ of the chaser, the runner imagines a goal location $\tilde{x}^\textsc{c}_\textsc{goal}$ at random and then uses the RRT planner to a sample a random future trajectory $\tilde{x}^\textsc{c}_{t:T}$. Since this model is not conditioned in any way, it returns weight 1.

\section{Planning as Inference Formulation}

The Chaser-Runner model performs two levels of nested inference. At the episode level, we infer the next time point $x^\textsc{c}_t$, conditioning on expected future detections. In order to evaluate this likelihood, we simulate runner trajectories that are conditioned to avoid future detections. We will perform inference using a nested importance sampling scheme \cite{naesseth2015nested}, which is a generalization of importance sampling in which weighted samples at one level in the model are used as proposals at other levels in the model. Note that nested importance sampling is \emph{not} a form of nested Monte Carlo estimation as discussed in \citet{rainforth2018nesting} and \citet{rainforth2018nestingb}. We discuss the distinctions between the two methods below.

We implement conditioning using a planning-as-inference formulation \cite{toussaint06,vandemeent2016black-box}. 
In planning-as-inference problems, a target density $\pi(x) = \gamma(x) / Z$ is defined in terms of a prior over trajectories $p(x)$ and the exponent of a utility or reward $R(x)$
\begin{equation}
    \gamma(x) = \exp(R(x)) p(x).
\end{equation}
The normalizing constant $Z = \mathbb{E}[\exp(R(x))]$ is sometimes referred to as the desirability \cite{todorov2009efficient}. 

The Chaser-Runner model in Algorithm~\ref{alg:chaser-runner-programs} defines a sequence of unnormalized densities 
\begin{align*}
    &\gamma_t(x^\textsc{c}_{t:T}, \tilde{x}^\textsc{c}_{t:T}, x^\textsc{r}_{1:T} \mid x^\textsc{r}_{t-1}) =
    \\
    &~~~~
    \exp[\alpha (T^\textsc{c}_\textsc{vis} \!-\! T^\textsc{r}_\textsc{vis} )] 
    \:
    p(x^\textsc{c}_{t:T} | x^\textsc{c}_{t-1})
    \:
    p(\tilde{x}^\textsc{c}_{t:T} | x^\textsc{c}_{t-1})
    \:
    p(x^\textsc{c}_{1:T}).
\end{align*}
In this density, the reward $\alpha (T^\textsc{c}_\textsc{visible} - T^\textsc{r}_\textsc{visible})$ depends on the \emph{difference} between the number of time points during which the chaser expects that the runner will be visible, and the number of time points during which the runner expects to be visible based the imagined chaser trajectory (which reflects a more naive model of the chaser). In other words, the the chaser aims to identify trajectories that will result in likely detections of the runner, under the assumption that the runner will avoid trajectories where detection is likely given a naive chaser model. 

\section{Nested Importance Sampling}

We can perform inference in the chaser-runner model using Monte Carlo methods for probabilistic programs. Algorithm~\ref{alg:chaser-runner-programs} defines an importance sampling scheme. At each time $t$, we sample $x^\textsc{c}_t \sim \pi_t(x^\textsc{c}_t \mid x^\textsc{c}_{t-1})$ from the marginal of the target density above. To do so, we sample $K$ particles from the \textsc{chaser} model (line 6). For each sample, in the \textsc{chaser}, we draw $L$ samples from the \textsc{runner} model (line 15). We then perform resampling to select $K$ of the resulting $K \cdot L$ particles with, which corresponds to performing SMC sampling within the \textsc{episode} model (lines 8-9). 


When $L=1$, this sampling scheme reduces to standard SMC inference for probabilistic programs \cite{wood-aistats-2014}. When $L>1$ it can be understood as a a form of nested importance sampling \cite{naesseth2015nested}. Note that in this sampling scheme, each of the $L$ samples corresponds to a \emph{different} runner trajectory $x^{\textsc{r},k,l}_{1:T}$, but that the reward for this trajectory is evaluated relative to the \emph{same} past $x^{\textsc{c},k}_{1:t-1}$ and imagined future $x^{\textsc{c},k}_{t:T}$ trajectory for the chaser. 

As noted above, nested importance sampling is not the same as nested Monte Carlo estimation. In nested Monte Carlo problems, we compute an expectation of the form $\mathbb{E}[f(y, \mathbb{E}[g(y, z)])]$, which is to say that we compute an expectation in which, for each sample $y$, we need to compute an expected value by marginalizing over samples $z$. In the chaser-runner problem, we would obtain a nested Monte Carlo problem if we defined the weight
\[
    w^k_t 
    = 
    \exp
    \left[
        \hat{R}\big(
            x^{\textsc{c}, k}_{1:T}
            \big)
    \right],
\]
by averaging the reward over chaser trajectories
\[
    \hat{R}\big(
        x^{\textsc{c}, k}_{1:T}
    \big)
    =
    \frac{1}{L}
    \sum_{l=1}^L
    R
    \big(
        x^{\textsc{c}, k}_{1:T},
        x^{\textsc{r}, k, l}_{1:T}
    \big)
    .
\]
In nested importance sampling, we select a particle $x^{\textsc{c}, k}_{1:T}$ according to the average weight 
\[
    w^k_t 
    = 
    \frac{1}{L}    
    \sum_{l=1}^L
    \exp 
    \left[
    R
    \big(
        x^{\textsc{c}, k}_{1:T},
        x^{\textsc{r}, k, l}_{1:T}
    \big)
    \right].
\]
This is sometimes referred to as nested conditioning, in the context of probabilistic programming systems \cite{rainforth2018nestingb}.
For any choice of $L$, this is a valid importance sampling scheme in which the importance weight provides an unbiased estimate of the normalizing constant.

The approach in Algorithm~\ref{alg:chaser-runner-programs} differs subtly from standard nested importance sampling approaches. Nested importance sampling was introduced as a means of reasoning about proposals in importance sampling that are themselves generated by means of another importance sampling algorithm. Concretely, suppose that we have an importance sampling mechanism that targets an unnormalized proposal density $x, w^\textsc{p} \eval \gamma^\textsc{p}(x)$ and wish to define an importance sampler that for an unnormalized target density $x, w^\textsc{t} \eval \gamma^\textsc{t}(x)$, then we can do so by defining the importance weight
\begin{align*}
    w^\textsc{t} 
    &= 
    \frac{\gamma^\textsc{t}(x) w^\textsc{p}}
         {\gamma^\textsc{p}(x)},
    &     
    x, w^\textsc{p} 
    &\eval \gamma^\textsc{p}(x).
\end{align*}
A corollary of this identity is that we may compose importance samplers to sample different subsets of variables in any generative model, e.g. we could propose using two importance samplers 
\begin{align*}
    x_2, w_2^\textsc{p} 
    &\eval 
    \gamma^\textsc{p}(x_2 \mid x_1)
    &
    x_1, w_1^\textsc{p}
    &\eval
    \gamma^\textsc{p}(x_1).
\end{align*}
and define the importance weight
\begin{align*}
    w^\textsc{t} 
    &= 
    \frac{\gamma^\textsc{t}(x_1, x_2) \: w_1^\textsc{p} \: w_2^\textsc{p}}
         {\gamma^\textsc{p}(x_2 \mid x_1) \gamma^\textsc{p}(x_1)}.
\end{align*}
In the sampling scheme Algorithm~\ref{alg:chaser-runner-programs}, we sequentially sample moves $x^\textsc{c}_t, w^\textsc{c}_t \eval \gamma_t(x^\textsc{c}_t \mid x^\textsc{c}_{t-1})$ from an importance sampler which has the same unnormalized density as the target, which can be understood as as special case of nested importance sampling in which $\gamma^\textsc{t}(x) = \gamma^\textsc{p}(x)$.
\vspace{-0.25em}
\section{Experiments}
\vspace{-0.25em}

We carry out three categories of experiments: 1) \textit{trajectory visualization experiments}, in which we qualitatively evaluate what forms of rational behavior arise in our model depending on conditioning, 2) \textit{detection rate experiments}, which test to what extent a more accurate model of a runner enables the chaser to detect the runner most often, and 3) \textit{sample budget experiments}, which serve to evaluate the trade-offs in allocating our sample budget across different levels of nesting in the model.

\begin{figure*}
\begin{center}
\centerline{\includegraphics[width=2.0\columnwidth]{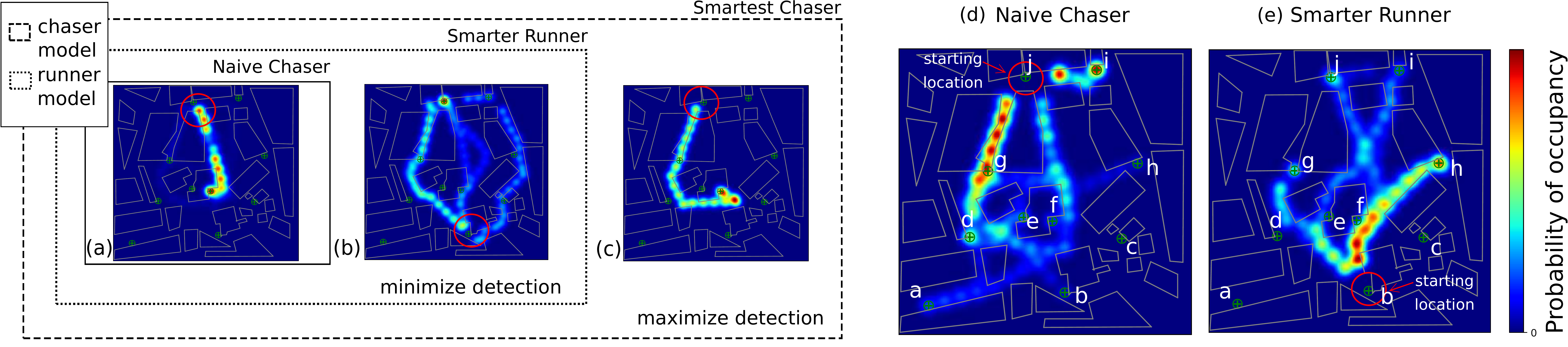}}
\caption{ \textbf{Chaser and runner trajectories} in the innermost, middlemost, and outermost models where locations circled in red are the starting locations for each agent. We show posterior distributions of $L$ \textit{runner} and \textit{naive chaser} paths, when $(K,L) = 128, 16$ for a single resampled sample $k$. (a)-(c) show posterior distributions over paths after running importance sampling where we condition the start and goal locations for each agent. Figures (d)-(e) show posterior paths after we only condition the start locations for the agents . 
}
\label{fig:progression}
\end{center}
\vspace{-1em}
\end{figure*} 

\begin{figure}[!t]
\begin{center}
\begin{minipage}[t]{0.4\linewidth}
\centerline{\textsf{\footnotesize Naive Runner}}
\end{minipage}
\begin{minipage}[t]{0.4\linewidth}
\centerline{\textsf{\footnotesize Smarter Runner}}
\end{minipage}

\centerline{\includegraphics[width=0.9\columnwidth]{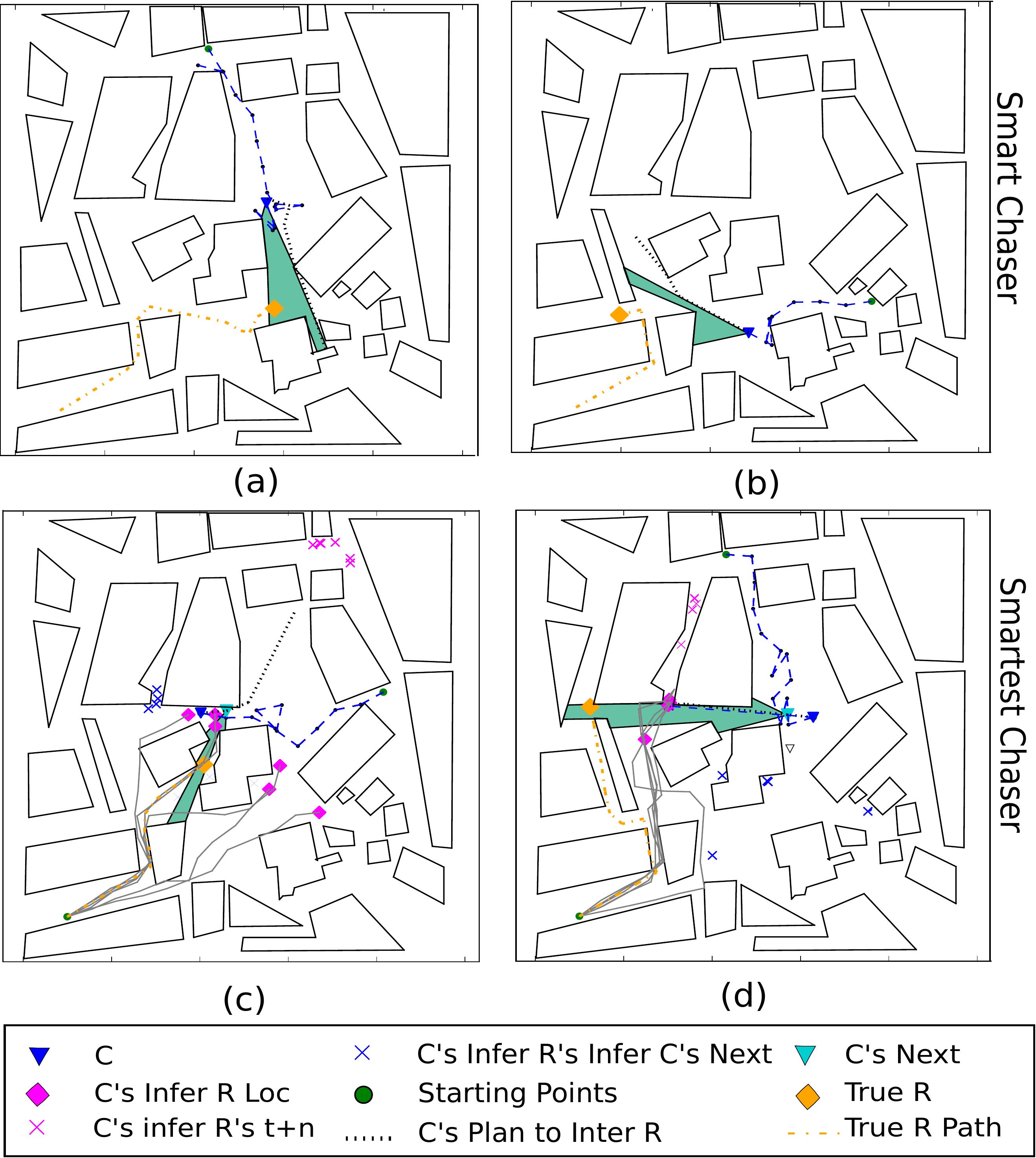}}
\caption{(a) Smart chaser playing against a naive runner, where the chaser anticipates the the intersection point and heads in the correct direction to detect the runner. (b) Smart chaser playing against a smarter runner. (c) Smartest chaser against the naive runner. (d) Smartest Chaser infers the runner locations to be more hidden, avoiding the center of the map. The chaser successfully detects the smarter runner.} 
\label{fig:exps}
\end{center}
\vspace{-1em}
\end{figure} 

\vspace{-0.25em}
\subsection{Visualization of Trajectories}
\vspace{-0.25em}

Before carrying out a more quantitative evaluation of the chaser-runner model, we visualize sampled trajectories to show how nested inference converges empirically to rational behavior at each level of the model. 
We begin by considering a simplified scenario in which we assume fixed start and goal locations. These locations are known to both the chaser and the runner, which means that the chaser and runner do not have to perform inference over possible goal locations.
Figure~\ref{fig:progression} (a) shows a heat map of naive chaser paths in the innermost model, which are conditioned on the start and goal locations. In Figure~\ref{fig:progression} (b), we show a heat map of runner paths, in which the runner travels in the opposite direction along the same two locations. We observe that the runner avoids direct routes so as to minimize chance of detection. In Figure~\ref{fig:progression} (c) we show a heat map of chaser trajectories in the outermost model, which shows that the chaser selects paths that are likely to lead to interception of the runner. Together, Figures~\ref{fig:progression} (a)-(c) demonstrate how our Chaser-Runner model can perform planning conditioned on start and end locations.

In Figure~\ref{fig:progression} (d)-(e), we visualize $L$ \textit{naive} chaser and runner paths from a single $k$ sample (sampled proportionally to importance weights) at times step $t\ge3$. 
The runner paths in Figure~\ref{fig:progression} (e) once again avoid detection relative to the naive chaser paths in Figure~\ref{fig:progression} (d). Although the naive chaser travels directly toward goal locations from the upper end of the map, in this particular $k$ sample, the naive chaser most often remains on the left side of the map. This results in the runner traveling through the center of city to minimize probability of detection, but more often planning toward location \textbf{h}.  This is a case where the RRT planner provides the runner with  a shorter and direct plan to minimize detection from the chaser.


\vspace{-0.25em}
\subsection{Detection Experiments}
\vspace{-0.25em}

To evaluate the influence of nested modeling on resulting plans, we compare detection rates in the full chaser-runner model to detection rates in three simplified models. We run simulations using two types of runners. We refer to the runner from the full model as the \emph{smarter runner}, and also consider a \textit{naive runner} which samples from the RRT planner in the same manner as the naive chaser. We similarly consider two chaser models. We refer to the chaser from the full model as the \emph{smartest chaser}. We additionally consider a simplified model in which the chaser assumes a naive runner, which we refer to as a \emph{smart chaser}. 

These two runner and chaser models together yield 4 modeling scenarios. Table~\ref{table:detect_rates} shows the average detection rate over 50 restarts for each scenario. Figure~\ref{fig:exps} shows illustrative trajectories. In this figure, `C' stands for Chaser; `R' stands for runner.  The blue triangle represents the chaser's true, current location. 
Blue dashed lines represent the past chaser trajectory whereas crosses mark future locations imagined by the runner. 
Magenta diamonds represent samples of inferred runner locations;
magenta crosses represent inferred future runner trajectories.

\paragraph{1. Naive Runner, Smart Chaser.} A smart chaser can reliably intercept a naive runner. 
Figure~\ref{fig:exps} (a) illustrates a successful detection. We observe that the chaser typically navigates to the center of the map. Since the shortest path between most points crosses the center of the map, this allows the chaser to intercept the runner with high probability.

\paragraph{2. Smarter Runner, Smart Chaser.} When we increase the model complexity of the runner, the detection probability decreases. 
Figure~\ref{fig:exps} (b) illustrates a prototypical result.  The \emph{smarter runner} expects the chaser to remain in the center of the map, as it is trying to head off a naive agent, and successfully avoids the center of the map. In Figure~\ref{fig:exps} (b), the runner is seen swerving sharply left taking a longer path around the perimeter of the city to reach its goal. As a result, the chaser is unable to find the runner for the rest of the simulation. The average detection rate is 0.36, which means that a smarter runner is able to avoid a misinformed chaser in most episodes.
\vspace{-0.25em}
\paragraph{3. Naive Runner, Smartest Chaser.} In this experiment, the chaser assumes a smarter runner, even though the runner's behavior is in fact naive.  Figure~\ref{fig:exps} (c) illustrates a prototypical result. Here, the multimodality of the model's inferences is apparent: the chaser predicts two possible modes where the runner could be (clusters of magenta triangles), but assigns more probability mass to the upper (correct) cluster; the result is that the chaser plans a path to that location, which results in a detection. As it turns out, this model variant results yields a detection rate of 0.98, which is the same as that of in scenario 1, where the chaser has an accurate model of the naive runner.

\textbf{4. Smarter Runner, Smartest Chaser.} Figure~\ref{fig:exps} (d) shows a prototypical result from the full chaser-runner model, which results in a successful detection. The chaser anticipates that the runner will avoid highly visible areas of the map and travel through alley ways and around the city. 

This experiment yielded a detection rate of 0.56, which is significantly higher than the detection rate of 0.36 in experiment 2. 

\begin{table}[!t]
\centering
\begin{tabular}{r|cc}
    \toprule
    & Naive Runner & Smarter Runner \\
    \midrule
    Smart Chaser & $(49/50)=0.98$ & $(18/50)=0.36$ \\
    Smartest Chaser & $(49/50)=0.98$ & $(28/50)=0.56$ \\
    \bottomrule
\end{tabular}
\caption{Detection Rates for Agent Model Variants}
\label{table:detect_rates}
\end{table}


\begin{figure}[!t]
\begin{center}
\centerline{
\rotatebox{90}{\hspace{6em}\scriptsize\textsf{ESS/K}\hspace{7em}\scriptsize\textsf{log $\mathsf{\bar{Z}}$}}
\hspace{-0.2em}\includegraphics[width=0.95\columnwidth]{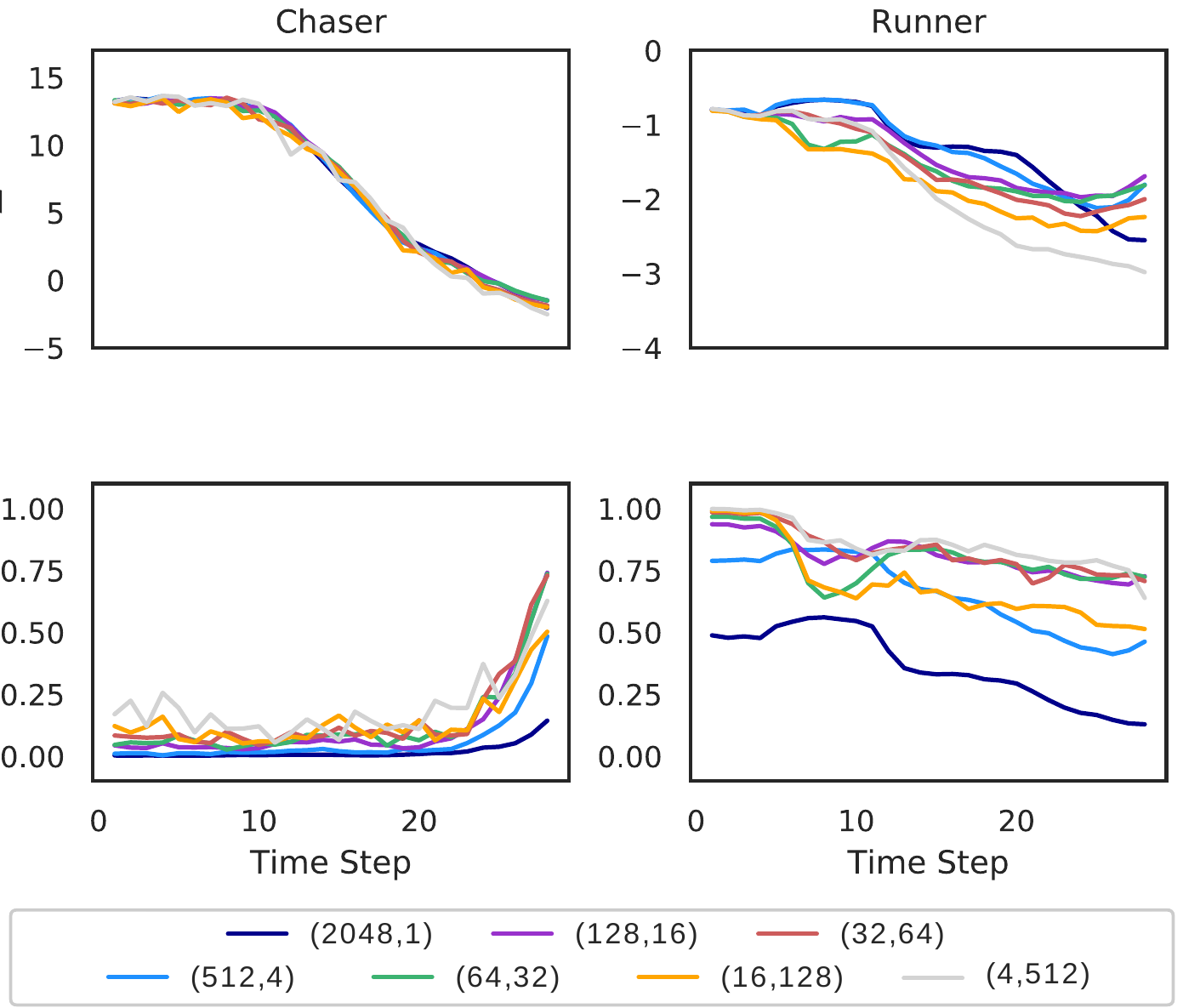}}
\caption{ Log mean log weights, $\log \bar{Z}$, and Fractional ESS (normalized by $K$) as a Function of Time for each sample budget. \textbf{Top Row:}  $\log \bar{Z}^\textsc{R}$ for the middlemost model (left), and $\log \bar{Z}^\textsc{C}$ the outermost model, (right). \textbf{Bottom Row:} The fractional ESS for each varying $K$ and $L$.  }
\label{fig:log_means}
\end{center}
\vskip -0.4in
\end{figure}

\textbf{Discussion.} These 4 scenarios illustrate that 
when the runner reasons more deeply, he evades more effectively; Conversely when the chaser reasons more deeply, he intercepts more effectively. Furthermore, we show that a single, unified inference algorithm can uncover a wide variety of intuitive, rational behaviors for both the runner and the chaser. 

\subsection{Sample Budget Experiments}

\begin{figure}[!t]
\begin{center}
\centerline{\hspace{-0.2em}\includegraphics[width=1.05\columnwidth]{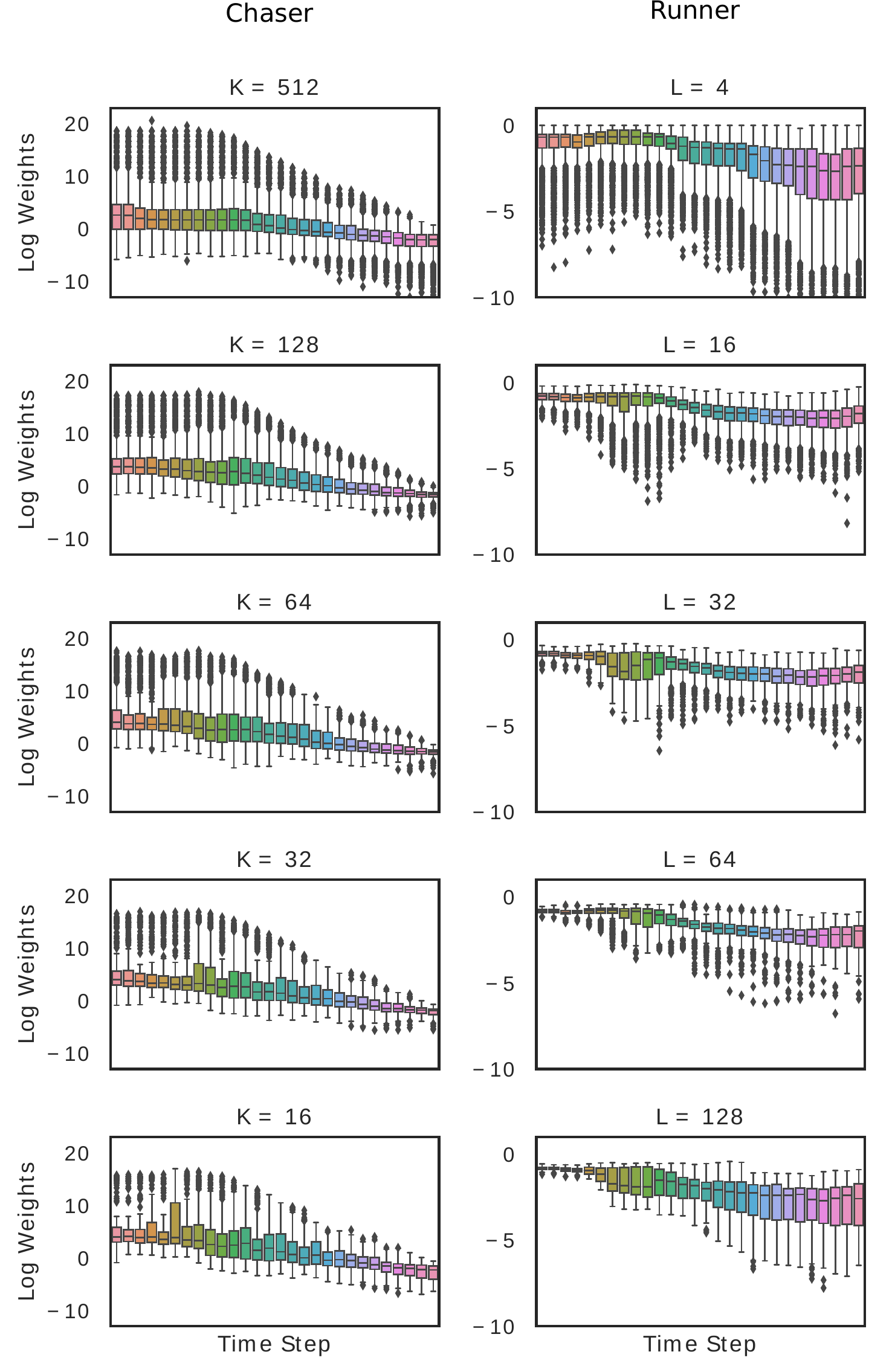}}
\caption{Box plots showing quantiles of log weights for the runner (left) and chaser (right) at each time step in the simulation for varying $K$ and $L$.}
\label{fig:box_plots}
\end{center}
\end{figure}

To evaluate how the allocation of computational resources to different levels of the model affects the variance of our importance sampling estimator, we carry out experiments in which we set $K$ and $L$ to
\[
    (K,L) = (2048,1), (512, 4), (128, 16), \ldots (4,512)
\]
This fixes the total computation budget to $K L = 2048$ samples, which allows us to assess how many samples from the runner are needed to effectively evaluate utilities in the chaser model.

In this experiment, we perform $R=10$ independent episode restarts for $7$ combinations of $(K,L)$ values. For each episode we compute $K$ chaser trajectories and $K \cdot L$ runner trajectories for  $T= 28$ time steps. In other words, we compute $7 \cdot R \cdot T \cdot K \cdot L$ runner trajectories $w^{\textsc{r}}$, which corresponds to just over 4 million calls to the RRT planner.

In Figure~\ref{fig:log_means} (top row), we show the log mean log weights for the chaser (left) and runner (right) at each time point $t$ 
\begin{align*}
    \log
    \bar{Z}^\textsc{c}_t 
    &=\frac{1}{R}
    \sum_{r=1}^R 
    \log 
    \left(
    \frac{1}{K}
    \sum_{k=1}^K
    \sum_{l=1}^L
    w^{\textsc{c},k,l}_t
    w^{\textsc{r},k,l}_t
    \right),
    \\
    \log
    \bar{Z}^\textsc{r}_t 
    &=\frac{1}{R}
    \sum_{r=1}^R 
    \log 
    \left(
    \frac{1}{KL}
    \sum_{k=1}^K
    \sum_{l=1}^L
    w^{\textsc{r},k,l}_t
    \right).
\end{align*}
For each sample budget, $\bar{Z}^\textsc{c}_t$ decreases (left) as a function of time while $\bar{Z}^\textsc{r}_t$ remain relatively stable independent of time (right). The decrease in $\bar{Z}^\textsc{c}_t$ is to be expected, given that the probability of intercepting the runner decreases as we approach the end of the episode.

To evaluate the weight variance at each time step, we compute the effective sample size (ESS), which for a set of $K \cdot L$ weights $\{w^{k,l}\}$ is defined as 
\[
    \textstyle \text{ESS} 
    = 
    \Big(
        \sum_{k=1}^K \sum_{l=1}^L w^{k,l} 
    \Big)^2 
    / 
    \Big( 
        \sum_{k=1}^K \sum_{l=1}^L (w^{k,l})^2
    \Big).
\]
Figure~\ref{fig:log_means} (bottom row), shows the fractional ESS 
(normalized by $K \cdot R$) 
as a function of time for each sample budget. The effective sample size for the chaser weights increases over the course of the episode, reflecting that inference becomes easier owing to the previously mentioned conclusion of progressively decreasing runner detection probabilities as we reach the end of the episode.

Figure~\ref{fig:box_plots} shows quantiles with respect to restarts for log mean weights, which further confirms the trend in Figure~\ref{fig:log_means}. We show higher median log weights and less outliers as K decreases and L increases,  $(K,L)=(16,128)$ results show that computed log weights are less robust when we draw a smaller number of samples from the outermost model. 

\vskip -0.1in
\section{Conclusion and Future Work}

In this paper, we have introduced a high-uncertainty variant of pursuit-evasion games where agents are required to reason about other agents' reasoning in order to accomplish their respective goals of pursing or evading other agents. 
We develop a planning-as-inference framework that enables us to perform online planning by way of nested simulations.
By formulating our models as probabilistic programs, we can incorporate semi-realistic deterministic primitives such as path planners and field-of-view calculations. 

To our knowledge, nesting importance sampling methods have not previously been applied to online planning in  multi-agent systems. Relative to existing approaches that model theory of mind with nested probabilistic programs, our work is the first to reason about agents in a time-dependent manner by repeatedly making inferences at each step of a simulation.
We empirically demonstrate that nested Bayesian reasoning leads to rational behaviors and that increasing model complexity to incorporate reasoning about reasoning outperforms non-nested models. 
Finally, our empirical evaluations indicate that nested reasoning results in lower-variance estimates of expected utility.


An advantage of probabilistic programming approaches is their compositionality.  While
here we assume knowledge of a high-level map, our framework could be
applied to a joint model that blends high-level reasoning with
low-level perception.  In such models, inferences in theory of
mind models could go beyond goals and paths, and could serve to
infer (for example) the existence of objects or other agents seen by
the runner, but not by the chaser. A future line of research is how to enable such integrated models via inference meta-programming architectures.



\bibliography{icml2019-tomsmc}
\bibliographystyle{icml2019}

\end{document}